\documentclass{article}
\usepackage{spconf,amsmath,graphicx,hyperref}
\hypersetup{colorlinks=false, pdfborder={0 0 0}}
\usepackage{tabularray}

\usepackage{times}
\usepackage{helvet}
\usepackage{courier}
\usepackage{xcolor}

\usepackage{booktabs}       
\usepackage{multirow}       
\usepackage{graphicx}       
\usepackage{array}         
\usepackage{makecell}       
\usepackage{xcolor}       
\usepackage{amssymb}       
\usepackage{amsmath} 

\usepackage{subcaption}

\title{GAMMA: Generalizable AI-Generated Image Detection via Multi-Task and Manipulation-Augmented Supervision}

\name{\shortstack{Haozhen Yan$^{1}$,
      Yan Hong$^{2}$,
      Suning Lang$^{1}$,
      Jiahui Zhan$^{1}$,
      Yikun Ji$^{1}$,\\
      Yujie Gao$^{1}$,
      Huijia Zhu$^{2}$,
      Jun Lan$^{2}$\sthanks{Corresponding authors.},
      Jianfu Zhang$^{1}$\footnotemark[1]}}
\address{$^{1}$ Shanghai Jiao Tong University, $^{2}$ Ant Group
\\ \small orion810@sjtu.edu.cn, lanjun\_yelan@163.com, c.sis@sjtu.edu.cn
}

\begin{document}
\ninept
\maketitle
\begin{abstract}
With the rapid advancement of generative models, detecting AI-generated images has become increasingly challenging. Although existing detectors achieve strong results on in-distribution samples, their generalization to unseen models remains limited, largely due to reliance on generation-specific biases such as image degradations, format inconsistencies, and dataset or model priors. To address these limitations, we propose \textbf{GAMMA}, a novel training framework that improves detector generalization through manipulated image augmentation and multi-task supervision. Manipulated images are introduced to alleviate generation-specific bias: inpainting maintains semantic fidelity but may introduce structural bias, whereas copy-move and splicing are model-agnostic and preserve structural plausibility while altering semantics. Their complementarity yields diverse and balanced samples that enhance robustness across various domains. A multi-task supervision scheme integrates pixel-level segmentation with image-level classification, where segmentation provides fine-grained analysis of local regions that supports more reliable global predictions. Furthermore, a reverse cross-attention mechanism enables segmentation to guide and refine biased representations in the classification branch. Extensive experiments demonstrate that GAMMA achieves state-of-the-art generalization on the GenImage benchmark, surpassing the prior best by 5.8\%, while maintaining strong performance on emerging models such as GPT-4o.
\end{abstract}

\section{Introduction}
\label{sec:intro}

The rapid progress of generative models~\cite{SD_2022,gpt4oimage_2025} has reshaped the landscape of synthetic content creation.
With pretrained architectures and prompt-based interfaces, users can now generate high-fidelity images without any expertise.
While this accessibility has greatly expanded creative possibilities, it also raises concerns regarding misuse, authenticity, and intellectual property~\cite{Epstein2023art}.
In practice, detectors aim to identify AI-generated images solely from visual cues.

Existing approaches achieve strong in-distribution performance but exhibit inherent limitations.
Low-level methods exploit frequency or microstructural artifacts~\cite{AIDE_2025}, yet are fragile under degradations and adversarial perturbations.
Reconstruction-based approaches leverage discrepancies with pretrained generators~\cite{DIRE_2023,AEROBLADE_2024}, yet are often hindered by compression artifacts and high computational cost.
Content-consistency strategies improve dataset quality and semantic alignment through inpainting~\cite{BFree_2025,COCO-Inpaint_2025,OpenSDI_2025}, but tend to overfit specific editing priors.
LLM-based methods~\cite{SIDA,FakeShield} provide interpretability at the expense of efficiency and accuracy.

Although these categories of methods address different aspects of the problem, they share a common weakness: their effectiveness drops sharply when encountering generative models unseen during training, \textit{i.e.}, generalizability remains fundamentally constrained.
This weakness, repeatedly documented in recent studies, has been attributed to several key factors: image storage formats~\cite{FakeJPEG_2025}, degradation from resizing or blurring~\cite{RandomCrop_2025}, and dataset bias~\cite{BFree_2025}.
We collectively refer to these issues as \textit{generation-specific bias}.
Instead of tackling the core detection objective, models often overfit to such superficial cues.
These challenges highlight that existing paradigms fall short of capturing the intrinsic nature of synthetic imagery, underscoring the need for a training framework that transcends binary classification and handcrafted artifacts. 
Detectors should instead focus on identifying the \textit{structural and semantic inconsistencies} inherent in synthetic content, rather than memorizing dataset-specific features or model-dependent noise.

\begin{figure*}[t]
	\centering
	\includegraphics[width=\linewidth]{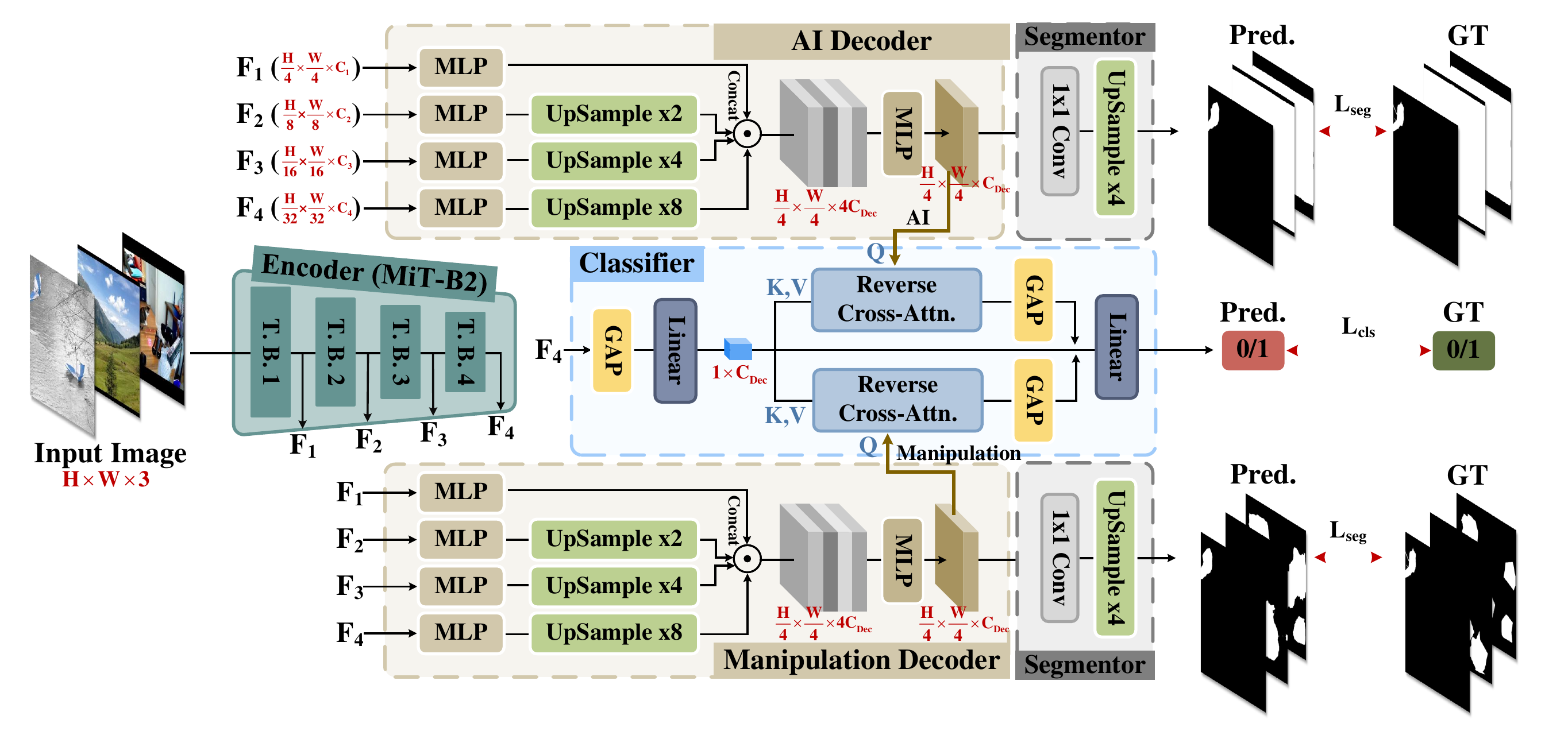}
        \vspace{-20pt}
	\caption{The Overview of GAMMA. Given an input image $\mathbf{I}\in\mathbb{R}^{H\times W\times 3}$, multi-scale features $\mathbf{F}_i$ are extracted via the MiT-B2 encoder and unified through MLP projection and upsampling. Two parallel decoders with identical architectures yield decoder outputs $\mathbf{F}_{\text{Dec}}^{AI}$ and $\mathbf{F}_{\text{Dec}}^{Ma}$, where $\mathbf{F}_{\text{Dec}} \in\mathbb{R}^{\frac{H}{4}\times\frac{W}{4}\times C_{\text{Dec}}}$; while the classifier predicts the global authenticity label with the aid of \textbf{Reverse Cross-Attention}, enabling the segmentation branches to rectify potentially biased classification responses (Fig.~\ref{fig:crossattn}).}
        \vspace{-16pt}
	\label{fig:pipeline}
\end{figure*}

\begin{figure}[t]
	\centering
	\includegraphics[width=\linewidth]{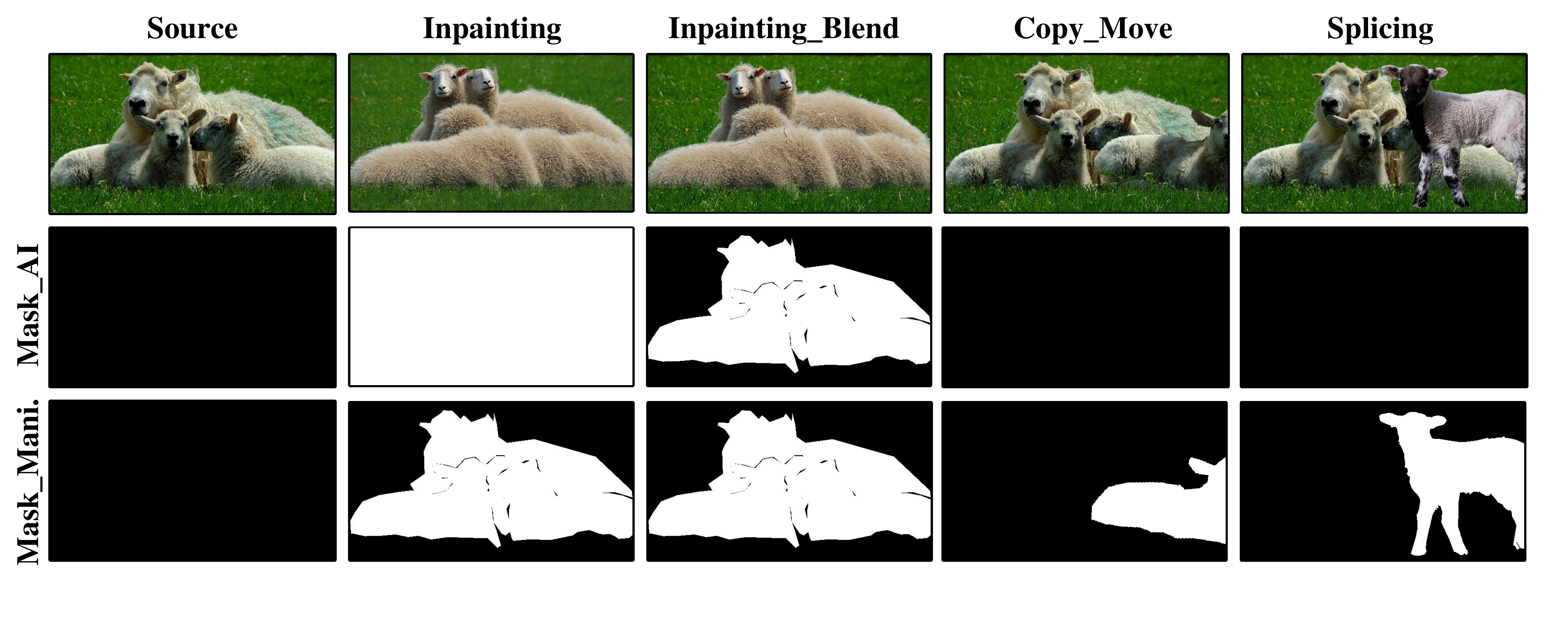}
        \vspace{-16pt}
	\caption{Different manipulation categories. In inpainting, unmasked regions also pass through diffusion and induce subtle pixel changes, yielding an all-white Mask\_AI.}
    \vspace{-16pt}
	\label{fig:manipulation_types}
\end{figure}

To address the limited generalizability of detectors, we propose GAMMA, a novel training framework that integrates manipulated image augmentation with multi-task supervision.
To mitigate generation-specific bias, we construct forgeries using inpainting, copy-move, and splicing operations (Fig.~\ref{fig:manipulation_types}). 
Inpainting preserves high-level semantics but may introduce structural anomalies due to generative model limitations. 
In contrast, copy-move and splicing rearrange real pixels in a model-agnostic manner, preserving structural plausibility while potentially disrupting local semantics.
These complementary manipulations yield diverse, content-aligned fake samples that prevent the detector from overfitting to artifacts of any single generative pipeline, and instead encourage it to capture structural and semantic inconsistencies that generalize across domains.
Building on these manipulations, we employ segmentation as auxiliary supervision for the binary classification task, enabling pixel-level attribution.
We further design a complementary segmentation labeling strategy: inpainting samples are labeled as AI-generated, ensuring semantic fidelity while exposing structural bias, whereas copy-move and splicing are labeled as real, preserving structure yet possibly altering semantics. This decoupling balances semantic- and structure-oriented supervision, mitigates generation-specific bias, and provides robust training signals for generalizable detection.
Furthermore, we introduce a reverse cross-attention mechanism that allows the segmentation branch to rectify potentially biased predictions in the classification branch.
We validate GAMMA on multiple datasets, achieving state-of-the-art generalization on GenImage~\cite{GenImage_2023} with a 5.8\% gain over the prior best, while remaining adaptable to new multimodal models such as GPT-4o~\cite{gpt4oimage_2025}.

\section{Methodology}
Our objective is to build a unified framework for classification, \textit{i.e.}, determining whether an image is AI-generated, while improving generalization through diverse manipulations and multi-task supervision.
To eliminate format-induced bias, all generated images originally stored in PNG are re-encoded into JPEG (quality 96, consistent with ImageNet~\cite{ImageNet_2009}). This step aligns synthetic images with real ones and prevents detectors from exploiting superficial compression artifacts~\cite{AEROBLADE_2024,FakeJPEG_2025}.
We then construct a training dataset comprising two types of manipulations (Fig.~\ref{fig:manipulation_types}): (1) semantic-aligned inpainting, which maintains high-level content while introducing localized edits, and (2) model-agnostic copy-move and splicing, which modify structural regions without introducing generation-specific bias. These manipulated samples are integrated into our proposed multi-task architecture (Fig.~\ref{fig:pipeline}) for training.

\subsection{Manipulation Augmentation}
To enrich training data while avoiding generation-specific bias, we incorporate diverse manipulations that either emphasize semantic fidelity or remain model-agnostic (Fig.~\ref{fig:manipulation_types}).\\
\noindent\textbf{Inpainting}. COCO-Inpaint~\cite{COCO-Inpaint_2025} offers diverse models, mask strategies, and text-guided or unguided modes. We adopt two variants: (1) content-based inpainting, where masked regions are restored purely from surrounding context, and (2) text-consistent inpainting, where the same prompt as the original image is used for reconstruction. Both variants maintain semantic alignment with the original content, yielding high-level consistency. 
To mitigate boundary artifacts, we further employ a blended version (COCO-Inpaint-Blended) that fuses unmasked regions from the original image with the inpainted output, improving realism while preserving semantic fidelity.
Due to limitations in generative models, inpainting may introduce structural anomalies or biases in the restored regions.\\
\noindent\textbf{Copy-Move and Splicing.} Following TampCOCO~\cite{TampCOCO_2022}, we also integrate copy-move and splicing operations, which rearrange real pixels without relying on generative models. These manipulations preserve structural plausibility but may disrupt local semantics. Crucially, they are model-agnostic and free from generation-specific artifacts, ensuring the detector does not overfit to any particular synthesis pipeline.

\begin{table}[t!]
\centering
\caption{Label configuration for training datasets. 0 indicates authentic, 1 indicates manipulated (Mani.) or AI-generated.}
\vspace{-8pt}
\label{tab:label_config}
\resizebox{0.8\linewidth}{!}{
\begin{tblr}{
  colspec={c c c c c c},     
  rows={c},                  
  cell{1}{1} = {r=3}{},      
  cell{1}{2} = {r=3}{},      
  cell{1}{3} = {c=4}{c},     
  cell{2}{3} = {c=2}{},      
  cell{2}{5} = {c=2}{},      
  cell{1-8}{1} = {l},        
  hline{1,8} = {-}{0.08em},  
  hline{2} = {3-6}{},        
  hline{3} = {3}{},          
  hline{3} = {4,6}{r},
  hline{3} = {5}{l},
  hline{4} = {-}{lr},        
}
Training Dataset & Cls. & Seg.  &       &       &       \\
                 &      & Mani. &       & AI    &       \\
                 &      & Bg     & Fg     & Bg    & Fg     \\
COCO~\cite{COCO_2014}                & 0 & 0 & 0 & 0 & 0 \\
COCO-Inpaint~\cite{COCO-Inpaint_2025}         & 1 & 0 & 1 & 1 & 1 \\
COCO-Inpaint-Blended & 1 & 0 & 1 & 0 & 1 \\
CopyMove\&Splicing~\cite{TampCOCO_2022}   & 0 & 0 & 1 & 0 & 0
\end{tblr}
}
\vspace{-16pt}
\end{table}

\subsection{Multi-Task Fine-Grained Supervision}
We employ segmentation as auxiliary supervision for the binary classification task, enabling pixel-level attribution. Such pixel-level guidance provides detailed analysis of local regions, which in turn supports more reliable image-level predictions.

\noindent \textbf{Label Configuration.}
Inpainting ensures semantic consistency but may introduce structural anomalies or biases due to limitations of generative models. Since our ultimate goal is to detect AI-generated images, inpainting constitutes the core supervision target. In contrast, copy-move and splicing preserve structural plausibility but may disrupt semantics. These operations are inherently model-agnostic and not directly aligned with the core detection objective, yet they provide valuable complementary signals. Together, the two manipulation types emphasize different aspects: semantic fidelity versus structural integrity, and jointly help mitigate generation-specific bias.
To leverage this complementarity, we adopt a multi-task strategy with two heads: a \textbf{Manipulation Segmentation Head}, which identifies any altered pixels, and an \textbf{AI Segmentation Head}, which localizes pixels synthesized by generative models. The former enforces localization of general manipulations, while the latter focuses on AI-specific artifacts such as stylistic or textural inconsistencies. This decoupled supervision allows the model to learn both broad manipulation priors and fine-grained generative cues, thereby improving generalization to unseen models. 
For the classification task, an image is defined as AI-generated if it contains regions predicted by the AI segmentation head. Accordingly, inpainting-based images are labeled as AI-generated, while copy-move and splicing samples, which duplicate only real pixels, are labeled as real (see Tab.~\ref{tab:label_config}). In this way, manipulation-based labels provide model-agnostic, structure-preserving samples, while inpainting supplies semantic-preserving samples that directly support the notion of AI-generated imagery.

\begin{figure}[t]
    \centering
    \includegraphics[width=\linewidth]{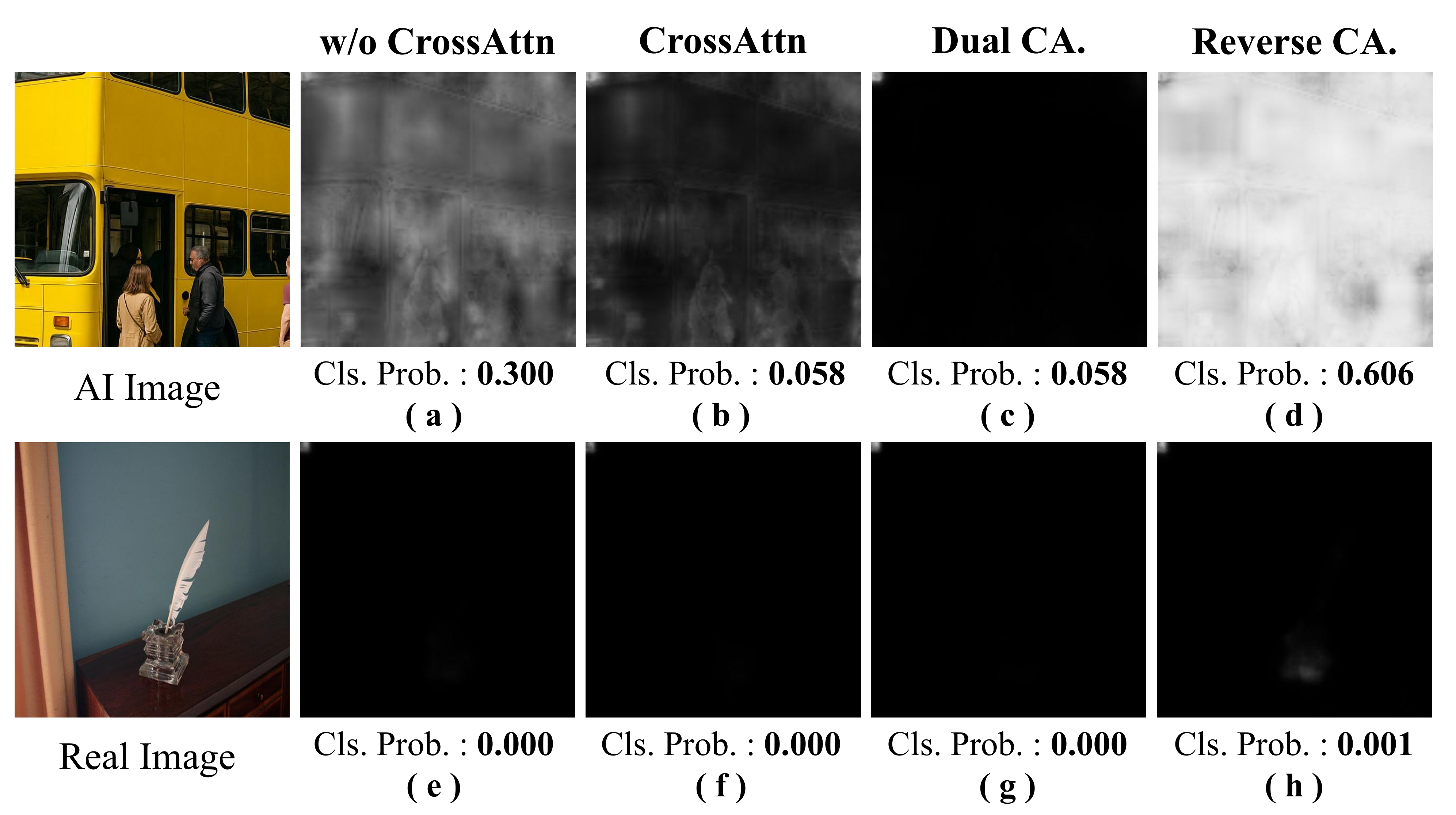}
    \vspace{-16pt}
    \caption{Impact of different Cross Attention (CA.) configurations on segmentation (AI Seg.) and classification performance. Cls. Prob. denotes the predicted probability of the image being AI-generated.}
    \label{fig:crossattn}
\end{figure}

\noindent \textbf{Model Architecture.}
We adopt SegFormer~\cite{SegFormer_2021} to predict manipulation and AI-generated masks. Given an input image \(\mathbf{I}\in\mathbb{R}^{H\times W\times 3}\), the ground truths are \(\mathbf{M}^{\mathrm{Ma}},\mathbf{M}^{\mathrm{AI}}\in\mathbb{R}^{H\times W}\). The encoder \(\mathbf{E}\) yields multi-stage features \(\mathbf{F}_i=\mathbf{E}(\mathbf{I})\in\mathbb{R}^{\frac{H}{r_i}\times\frac{W}{r_i}\times C_i}\) for \(i\in\{1,2,3,4\}\) with \(r_i\in\{4,8,16,32\}\).
The multi-scale features ${F_i}$ are projected to a common channel dimension, spatially aligned, and concatenated:
\begin{equation}
    \mathbf{F}_{\text{concat}} = \mathrm{Concat}\!\left(\mathrm{Up}_{s_i}\big(\mathrm{MLP}_i(\mathbf{F}_i)\big)\right)_{i=1}^4,
\end{equation}
where $s_i \in \{1, 2, 4, 8\}$ denotes the upsampling factor.
The concatenated features are refined by convolutional layers and an MLP to produce decoder features:
\begin{equation}
    \mathbf{F}_{\text{Dec}} = \mathrm{MLP}\big(\mathrm{Conv}(\mathbf{F}_{\text{concat}})\big)\in\mathbb{R}^{\frac{H}{4}\times\frac{W}{4}\times C_{\text{Dec}}}.
\end{equation}
AI and Manipulation decoders share the same architectural design, yielding $\mathbf{F}_{\text{Dec}}^{AI}$ and $\mathbf{F}_{\text{Dec}}^{Ma}$, which are each projected with a \(1\times1\) convolution and upsampled to produce the prediction map:
{
\begin{equation}
\hat{\mathbf{M}} = \mathrm{Up}_{4} \left( \mathrm{Conv}_{1 \times 1} \left( \mathbf{F}_{\mathrm{Dec}} \right) \right),
\hat{\mathbf{M}} \in \mathbb{R}^{H \times W}.
\end{equation}
}%
For classification, we apply global average pooling on $\mathbf{F}_{4}$ followed by a linear projection:
\begin{equation}
\mathbf{F}_{\text{Dec}}^{\text{Cls.}} = \mathrm{Linear}\big(\mathrm{GAP}(\mathbf{F}_{4})\big) \in \mathbb{R}^{1 \times C_{\text{Dec}}}.
\end{equation}
The full training objective is defined as:
\begin{equation}
    \mathcal{L}
=\alpha(\mathcal{L}_{\mathrm{Cls.}}+\mathcal{L}_{\mathrm{Seg}_{AI}})+\mathcal{{L}}_{\mathrm{Seg}_{Ma}},
\end{equation}
where $\alpha=2$ is a balancing hyper-parameter.

\begin{table}[t]
\centering
\caption{Generalization on GenImage (JPEG-96).}
\label{tab:genimage}
\begin{tblr}{
  colspec = {l c l c}, 
  hline{1-2,10-11} = {-}{},
  hline{1,11} = {-}{0.08em},
  column{1,3} = {halign=l},
  column{2,4} = {halign=c},
}
Method & Acc$\uparrow$ & Method & Acc$\uparrow$ \\
CNNDetect~\cite{Wang_2020_CVPR} & 51.3 & AntifakePrompt~\cite{Chang2023antifake} & 78.5 \\
DMID~\cite{Corvi_2023} & 79.0 & NPR~\cite{Tan2024rethinking} & 50.7 \\
LGrad~\cite{Tan2023learning} & 39.6 & FatFormer~\cite{Liu2024forgery} & 61.5 \\
UnivFD~\cite{Ojha2023towards} & 65.5 & FasterThanLies~\cite{Lanzino2024faster} & 77.0 \\
DeFake~\cite{Sha2023defake} & 73.7 & RINE~\cite{Koutlis2024leveraging} & 69.1 \\
DIRE~\cite{DIRE_2023} & 47.3 & AIDE~\cite{AIDE_2025} & 60.2 \\
LaDeDa~\cite{Cavia2024real} & 50.2 & C2P-CLIP~\cite{Tan2024c2p} & 75.5 \\
CoDE~\cite{Baraldi2024contrastive} & 71.7 & B-Free~\cite{BFree_2025} & 89.3 \\
\SetCell[c=4]{c} GAMMA (ours) : \, \textbf{95.1} \\
\end{tblr}
\end{table}

\begin{table}[t]
  \centering
  \caption{Generalization on other generative datasets.}
  \label{tab:other_datasets}
  \resizebox{\linewidth}{!}{
  \begin{tblr}{
    colspec = {l c c c c},
    hline{1,6} = {-}{0.08em},
    hline{2} = {-}{},
    row{1} = {font=\bfseries},
  }
    Dataset       & DMID & C2P-CLIP & B-Free & GAMMA \\
    SynthWildX~\cite{SynthWildX_2024}    & 48.3 & 58.2     & 95.6   & \textbf{99.1} \\
    OpenAI-4o~\cite{OpenAI-4o_2025}     & 49.8 & 49.8     & 53.5   & \textbf{79.4} \\
    GPT-ImgEval~\cite{GPT-ImgEval_2025}   & 50.5 & 49.6     & 57.5   & \textbf{84.1} \\
    ShareGPT-4o~\cite{ShareGPT-4o-Image_2025}   & 50.2 & 49.1     & 55.5   & \textbf{70.2} \\
  \end{tblr}
  }
\end{table}

\noindent \textbf{Reverse Cross Attention.}
When generalizing to novel samples, we observe that most existing detectors tend to misclassify photorealistic AI-generated images as real.
In contrast, segmentation heads can still capture localized manipulation cues in such cases due to their fine-grained spatial sensitivity (Fig.~\ref{fig:crossattn}(a)).
Notably, for authentic images, the segmentation predictions are generally consistent with the classification results (Fig.~\ref{fig:crossattn}(e-h)).
Motivated by this observation, we propose to enhance the interaction between the segmentation and classification branches by introducing a reverse cross-attention mechanism.
Specifically, the cross-attention is estimated as:
{\small
\begin{equation} 
\mathrm{Attention}(Q, K, V) = \mathrm{Softmax} \left( \frac{QK^\top}{\sqrt{d_\mathrm{head}}} \right) V,
\end{equation}}%
where we set $Q = \mathbf{F}_{\text{Dec}}^t$, with $t \in \{\text{Ma}, \mathrm{AI}\}$ and $K, V = \mathbf{F}_{\text{Dec}}^{\text{Cls.}}$.
The resulting attention is then residually added back to the classification features:
{\small
\begin{equation}
\mathbf{F}_{\text{Dec}}^{\text{Cls.}} = \mathbf{F}_{\text{Dec}}^{\text{Cls.}} + \sum_{t} \mathrm{GAP}(\mathrm{Attn}(\mathbf{F}_{\text{Dec}}^t, \mathbf{F}_{\text{Dec}}^{\text{Cls.}}, \mathbf{F}_{\text{Dec}}^{\text{Cls.}})).
\end{equation}}%
Intuitively, the reverse attention allows the segmentation head to actively reinforce informative cues from the classification representation, providing fine-grained spatial guidance.
This mechanism effectively enables the segmentation branch to rectify potentially biased classification responses (Fig.~\ref{fig:crossattn}(d)).
In contrast, the standard cross-attention, where the classifier serves as the query, may propagate the classification bias further (Fig.~\ref{fig:crossattn}(b)). Although dual cross-attention incorporates bidirectional attention into the classification feature, it introduces interference due to semantic inconsistency between the AI-generated and manipulation branches.

\begin{table}[t]
\centering
\caption{Ablation study on segmentation heads and Classification label. 0 indicates authentic, 1 indicates manipulated or AI-generated. none indicates only the COCO-Inpaint dataset is used for training.}
\begin{subtable}{0.48\linewidth}
\centering
\scalebox{0.82}{
\begin{tblr}{
  hline{1,6} = {-}{0.08em},
  hline{2} = {1}{r},
  hline{2} = {2}{lr},
}
Head                               & Acc$\uparrow$    \\
Cls.~                    & 91.0 \\
Cls. + ManiSeg           & 91.9 \\
Cls. + AISeg             & 92.3 \\
Cls. + ManiSeg +  AISeg  & 93.2 
\end{tblr}
}
\caption{Segmentation heads}
\label{tab:abla_seg_head}
\end{subtable}
\hfill
\begin{subtable}{0.48\linewidth}
\centering
\scalebox{0.8}{
\begin{tblr}{
  cells = {c},
  cell{1}{1} = {c=2}{},
  cell{1}{3} = {r=2}{},
  hline{1,8} = {-}{0.08em},
  hline{2} = {1}{l},
  hline{2} = {2}{r},
  hline{3} = {-}{lr},
}
Classification
  Label~ &        & Acc$\uparrow$    \\
Copy-Move               & Splicing &        \\
none                    & none   & 90.3 \\
1                       & 1      & 89.4 \\
0                       & 1      & 91.0 \\
1                       & 0      & 91.9 \\
0                       & 0      & 92.1 
\end{tblr}
}
\caption{Classification label}
\label{tab:abla_tampcoco_label}
\end{subtable}
\end{table}

\begin{table}[t]
\centering
\caption{Ablation study on manipulation diversity.}
\vspace{-8pt}
\label{tab:abla_dataset}
\begin{tblr}{
  hline{1,6} = {-}{0.08em},
  hline{2} = {-}{},
}
Training  Dataset                        & Acc$\uparrow$ \\
Inpainting                               & 90.6        \\
Inpainting + Blended                     & 90.9        \\
Inpainting + Copy-Move\&Splicing           & 92.1        \\
Inpainting + Blended + Copy-Move\&Splicing & 93.2        
\end{tblr}
\end{table}

\begin{table}[t]
\centering
\caption{Ablation study on CrossAttn and loss function weights.}
\vspace{-8pt}
\label{tab:abla}
\begin{subtable}{0.48\linewidth}
\centering
\begin{tblr}{
  hline{1,6} = {-}{0.08em},
  hline{2} = {-}{},
}
CrossAttn Config  & Acc$\uparrow$ \\
w/o CrossAttn     & 91.8        \\
CrossAttn         & 90.6        \\
Dual CrossAttn    & 90.8        \\
Reverse CrossAttn & 93.2        
\end{tblr}
\caption{CrossAttn}
\label{tab:abla_crossattn}
\end{subtable}
\hfill
\begin{subtable}{0.48\linewidth}
\centering
\begin{tblr}{
  colspec = {Q[c,0.12\linewidth] Q[c,0.18\linewidth] Q[c,0.20\linewidth] Q[c,0.12\linewidth]},
  vline{4} = {1-6}{},
  hline{1,6} = {-}{0.08em},
  hline{2} = {-}{},
}
$\mathcal{L}_{\mathrm{Cls.}}$ & $\mathcal{L}_{\mathrm{Seg}_{AI}}$ & $\mathcal{L}_{\mathrm{Seg}_{Ma}}$ & Acc$\uparrow$ \\
2 & 1 & 1 & 91.1 \\
2 & 3 & 1 & 89.4 \\
2 & 1 & 2 & 90.4 \\
2 & 2 & 1 & 93.2 
\end{tblr}
\caption{Loss weights}
\label{tab:loss}
\end{subtable}
\end{table}

\section{Experiments}

\subsection{Implementation Details}
We adopt SegFormer-B2~\cite{SegFormer_2021} with MiT-B2 ImageNet-pretrained weights as the backbone. The model is trained end-to-end with binary cross-entropy on NVIDIA A100 GPUs, using a batch size of 16, an initial learning rate of 1e-4, and early stopping with patience 5. 
To preserve forensic traces, RandomCrop ($512 \times 512$) is applied during training and CenterCrop during evaluation~\cite{RandomCrop_2025}, with padding and resizing for small images (smaller than $512 \times 512$).
Data augmentation follows~\cite{Wang_2020_CVPR}: JPEG quality 30--100, Gaussian blur (kernel size 1--19), and color jitter ($\pm 0.1$), each at 50\% probability.
Training converges in $\sim$6k steps ($\sim$24 GPU-hours) with early stopping.

\subsection{Generalization Evaluation}

GenImage~\cite{GenImage_2023} covers eight generative models with ImageNet-aligned prompts, and we evaluate on 1,000 generated and 1,000 authentic images per model. 
Tab.~\ref{tab:genimage} shows that GAMMA outperforms the state-of-the-art~\cite{BFree_2025} by \textbf{5.8\%} accuracy and achieves consistent results across subsets, highlighting strong generalizability.

We further benchmark the best-performing methods on four additional datasets, and the results are in Tab.~\ref{tab:other_datasets}.
Given the high fidelity of GPT-4o generated images, we evaluate on three related datasets~\cite{OpenAI-4o_2025,ShareGPT-4o-Image_2025,GPT-ImgEval_2025}, each paired with an equal number of ImageNet images.
GAMMA consistently outperforms prior methods on all four datasets and reaches 99.1\% accuracy on SynthWildX~\cite{SynthWildX_2024}, highlighting strong robustness across diverse distributions.
Unlike competing methods that often misclassify high-quality forgeries as real, GAMMA benefits from artifact-decoupling segmentation and reverse cross-attention, which jointly prevent overconfidence on realistic synthetic images.

\subsection{Ablation Studies}
We conduct ablation studies to evaluate the contribution of each component. The results reported in this section are averaged over all GenImage subsets and GPT-4o-related datasets.

\noindent \textbf{Segmentation Head.} As shown in Tab.~\ref{tab:abla_seg_head}, both Manipulation Segmentation (``ManiSeg'') and AI Segmentation (``AISeg'') individually improve classification accuracy, with AISeg yielding larger gains. Using both together achieves the best performance, confirming their complementarity.

\noindent \textbf{Classification Label.}
We analyze different label definitions on COCO-Inpaint in Tab.~\ref{tab:abla_tampcoco_label}. The baseline without TampCOCO reaches 90.3\%. Treating TampCOCO as real yields the best results, whereas labeling them as manipulated degrades performance, reflecting their non-generative nature and structural inconsistencies that improve robustness.

\noindent \textbf{Manipulation Diversity.}
Tab.~\ref{tab:abla_dataset} shows that combining multiple manipulation types consistently improves detection, confirming the benefit of diverse supervision.

\noindent \textbf{Cross Attention.}
As shown in Tab.~\ref{tab:abla_crossattn}, using classification features as Query degrades performance due to confirmation bias, while Reverse CrossAttn achieves the highest accuracy (93.2\%) by guiding classification with segmentation-derived cues. The results confirm that the segmentation-to-classification direction is most effective for attention.

\noindent \textbf{Loss Weights.} Tab.~\ref{tab:loss} indicates that appropriate weighting of AISeg loss improves robustness, whereas excessive weight impairs training and reduces overall accuracy.

\section{Conclusions}
Existing detectors generalize poorly due to reliance on generation-specific artifacts.
To address this, we introduce GAMMA, which reduces domain bias and enforces semantic alignment through diverse manipulations such as inpainting and semantics-preserving perturbations, multi-task supervision with dual segmentation–classification heads for fine-grained source attribution, and a reverse cross-attention module that allows segmentation to correct biased classifier features.
GAMMA achieves state-of-the-art out-of-distribution generalization on the GenImage benchmark, improving accuracy by 5.8\% and remaining robust to recent generative models such as GPT-4o.

\section{Acknowledgments}
This work was supported in part by the National Natural Science Foundation of China (Grant Nos. 62302295, 62595733, and 62561160155), the Shanghai Municipal Science and Technology Major Project (Grant No. 2021SHZDZX0102). This work was also supported by Ant Group.


\bibliographystyle{IEEEbib}
\bibliography{strings,refs}

\end{document}